\pdfoutput=1
\documentclass[11pt]{article}

\usepackage[final]{acl}

\usepackage{times}
\usepackage{latexsym}
\usepackage{multirow}
\usepackage{graphicx}
\usepackage{booktabs}
\usepackage{enumitem}
\usepackage{algorithm}
\usepackage{algorithmic}
\usepackage{amsmath}
\usepackage{bm}
\usepackage{colortbl}
\usepackage{subcaption}
\usepackage{tabularx}
\usepackage{pifont}
\usepackage{siunitx}     % 数字对齐
\usepackage[T1]{fontenc}
\usepackage[utf8]{inputenc}

\usepackage{microtype}

\usepackage{inconsolata}

\def\method{\textsc{RobustFT}}
\title{
{\includegraphics[height=0.5cm]{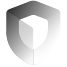}}
\method{}: Robust Supervised Fine-tuning for Large Language Models under Noisy Response
}

\makeatletter
\renewcommand{\@fnsymbol}[1]{^\dagger}
\makeatother
\author{
Junyu Luo\textsuperscript{\rm $\heartsuit$}, 
Xiao Luo\textsuperscript{\ding{171}}, 
Kaize Ding\textsuperscript{\rm $\diamondsuit$}, 
Jingyang Yuan\textsuperscript{\rm $\heartsuit$}, 
Zhiping Xiao\textsuperscript{\ding{168}},
Ming Zhang\textsuperscript{\rm $\heartsuit$} \\
{\textsuperscript{\rm $\heartsuit$} Peking University} 
\quad 
{\textsuperscript{\ding{171}} University of California, Los Angeles}
\\ 
{\textsuperscript{\rm $\diamondsuit$} Northwestern University}  
\quad
{\textsuperscript{\ding{168}} University of Washington}
\\
{\tt \small luojunyu@stu.pku.edu.cn, xiaoluo@cs.ucla.edu, kaize.ding@northwestern.edu}\\ {\tt \small patxiao@uw.edu, \{yuanjy, mzhang\_cs\}@pku.edu.cn}
}

\def\eg{\emph{e.g}.} 
\def\ie{\emph{i.e}.}

\definecolor{LightCyan}{rgb}{0.88,1,1}

\newcommand{\paratitle}[1]{\noindent\emph{\textbf{#1}}}
\usepackage{amsmath,amsfonts,bm}
\usepackage{algorithm,amsmath,amssymb,bm,amsthm}
\usepackage{algorithmic}
\usepackage{enumitem}

\newtheorem*{lemma*}{Lemma}

\def\eqref#1{equation~\ref{#1}}
\def\1{\bm{1}}

\DeclareMathAlphabet{\mathsfit}{\encodingdefault}{\sfdefault}{m}{sl}
\SetMathAlphabet{\mathsfit}{bold}{\encodingdefault}{\sfdefault}{bx}{n}

\def\gD{{\mathcal{D}}}

\def\gM{{\mathcal{M}}}
\def\gN{{\mathcal{N}}}

\DeclareMathOperator*{\argmin}{arg\,min}

\begin{document}
\maketitle
\begin{abstract}

Supervised fine-tuning (SFT) plays a crucial role in adapting large language models (LLMs) to specific domains or tasks. However, as demonstrated by empirical experiments, the collected data inevitably contains noise in practical applications, which poses significant challenges to model performance on downstream tasks. Therefore, there is an urgent need for a noise-robust SFT framework to enhance model capabilities in downstream tasks.
To address this challenge, we introduce a robust SFT framework (\method{}) that performs noise detection and relabeling on downstream task data. For noise identification, our approach employs a multi-expert collaborative system with inference-enhanced models to achieve superior noise detection. In the denoising phase, we utilize a context-enhanced strategy, which incorporates the most relevant and confident knowledge followed by careful assessment to generate reliable annotations.
Additionally, we introduce an effective data selection mechanism based on response entropy, ensuring only high-quality samples are retained for fine-tuning.
Extensive experiments conducted on multiple LLMs across five datasets demonstrate \method{}'s exceptional performance in noisy scenarios. Our code and data are publicly available.\footnote{\url{https://github.com/luo-junyu/RobustFT}}
% at \url{}.

\end{abstract}

\section{Introduction}

Supervised fine-tuning (SFT) has emerged as a critical technique for optimizing Large Language Models' (LLMs) capabilities, particularly in domain-specific tasks and adapting them to specific scenarios~\cite{minaee2024large,raffel2020exploring}. 
High-quality scenario-specific data plays a vital role in enhancing model performance on downstream tasks~\cite{xia2024less,jeong2024llm}.
While such data can be acquired through various means including human annotation ~\cite{li2024getting}, scenario-specific collection ~\cite{clark2019boolq}, and model-based self-labeling ~\cite{wang2024self}, these data sources inherently contain noise stemming from both human annotation errors and model hallucinations ~\cite{farquhar2024detecting}.

\begin{figure}[t]
    \centering
    \includegraphics[width=\linewidth]{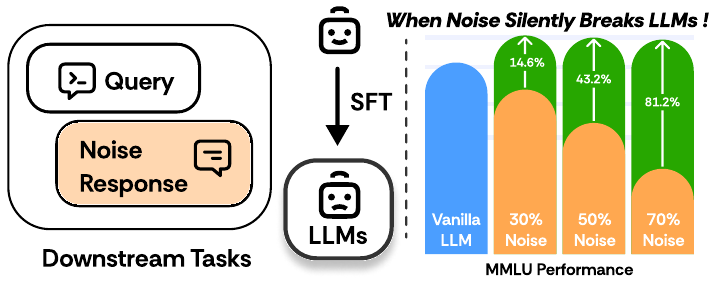}
    % \vspace{-2mm}
    \caption{Impact of noisy data on LLM performance during SFT. Increasing noise levels deteriorates model performance, highlighting the critical need for noise-robust fine-tuning approaches.}
    \label{fig:teaser}
    \vspace{-2mm}
\end{figure}

Data noise in scenario can have catastrophic effects on model performance. As shown in Figure~\ref{fig:teaser}, the MMLU~\cite{mmlu} evaluation results clearly demonstrate this degradation: as the proportion of noisy data increases, model accuracy shows a sharp decline. Specifically, with just $30\%$ noise in the training data, the model's performance deteriorates by $8.9\%$ compared to the vanilla LLM baseline. This performance degradation becomes increasingly severe as noise levels rise further. These findings underscore the critical importance and practical value of developing noise-robust fine-tuning frameworks for LLMs to maintain reliable downstream performance.
This motivates our central research question:
\begin{quote}
\vspace{-2mm}
\textit{Can LLMs detect inevitable noise and enhance data quality, to improve its performance on target tasks?}
\vspace{-2mm}
\end{quote}

The development of a noise-robust LLM fine-tuning framework encounters two major challenges. 
\textit{First}, direct noise detection through LLM predictions proves unreliable due to model hallucinations and overconfidence, as validated by our empirical studies in Section~\ref{sec:experiment}. 
\textit{Second}, while existing noise-robust methods work well for classification tasks with discrete label spaces~\cite{yuan2024hide,wang2023noise}, they are inadequate for LLM fine-tuning scenarios that require contextual and open-ended text generation. Traditional relabeling strategies not only fail to utilize valuable information in noisy generated responses. 
These challenges highlight the complexity of developing a framework that effectively leverages both model capabilities and data characteristics for robust noise detection and denoising in LLM fine-tuning.

In this paper, we propose \method{}~(Noise-robust LLM Supervised Fine-Tuning), a framework for effective adaptation in downstream scenarios with noisy data. 
At its core, \method{} introduces multi-view noise detection and denoising strategies. 
For noise detection, \method{} employs a collaborative multi-expert system, incorporating \textit{reasoning-enhanced} models to identify potentially noisy data effectively. 
For the identified noisy data, \method{} designs a denoising and data selection process. First, \method{} utilizes high-confidence data as contextual references for reliable relabeling of noisy samples. Subsequently, for both \textit{context-enhanced} and \textit{reasoning-enhanced} inference, \method{} employs \texttt{Review} Agent to examine and synthesize responses. Finally, by computing confidence scores based on model response entropy and excluding low-confidence samples, we obtain a denoised fine-tuning dataset that facilitates model adaptation to downstream tasks. 
Overall, by combining noise detection and denoising processes, \method{} effectively enhances the quality of the fine-tuning dataset while maximizing data utility. 
We validate \method{}'s effectiveness through extensive experiments across five datasets, spanning both general and domain-specific tasks with varying noise levels. 
Through comprehensive comparative analyses and ablation studies, we demonstrate the superiority of our approach.

Our contributions can be summarized as follows:
\begin{itemize}[leftmargin=*]
\vspace{-2mm}
\item \textbf{New Perspective.} We investigate the critical yet understudied challenge of noise-robust supervised fine-tuning for LLMs, which aligns more closely with real-world scenarios where noise is inevitable.
\vspace{-2mm}
\item \textbf{Principled Methodology.} We design a self-contained framework to leverage the intrinsic interactions between models and data for effective noise detection and denoising, eliminating dependencies on external models or resources.

\item \vspace{-2mm} \textbf{Superior Performance.} \method{} exhibits robust performance across diverse noise conditions, demonstrating significant improvements on three open-source LLMs across both general and domain-specific tasks, which validates its broad applicability and practical value.
\end{itemize}

% \clearpage

\section{Preliminary}

\begin{figure*}[t]
    \centering
    \includegraphics[width=\textwidth]{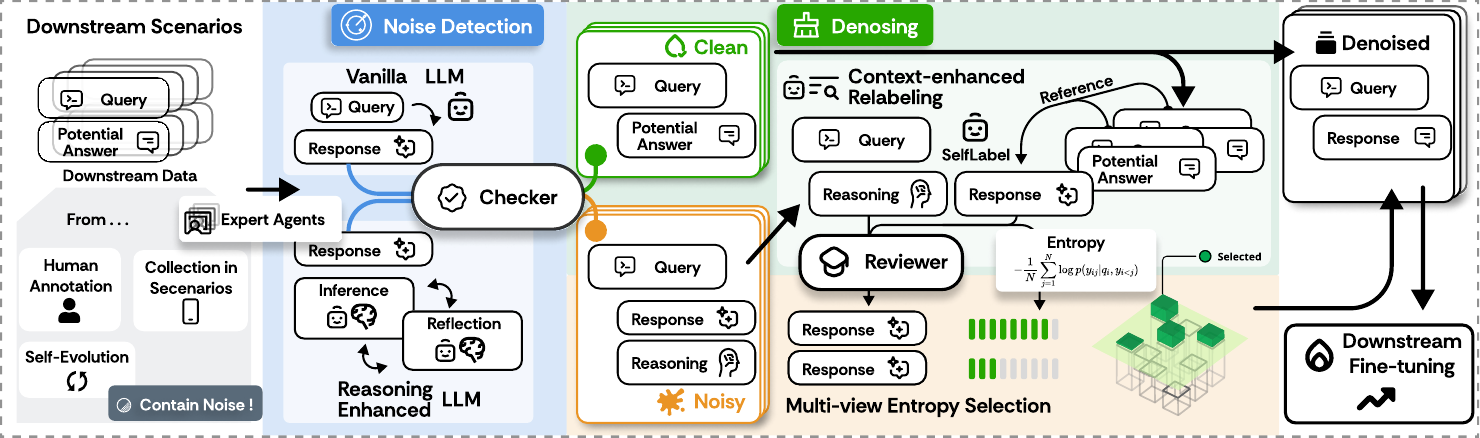}
    \caption{Overview of \method{}. Our \method{} enhances model performance through a two-stage noise \textit{detection-and-denoising} framework, leveraging collaborative learning among expert LLMs for noise detection and context-enhanced reasoning for data denoising, ultimately enabling robust downstream fine-tuning.}
    \label{fig:framework}
    \vspace{-2mm}
\end{figure*}

\subsection{Real-world Challenge}

In practical applications of Large Language Models (LLMs), our objective extends beyond enhancing their general capabilities to improving their performance on downstream tasks. To achieve this, we utilize Supervised Fine-Tuning (SFT) to optimize an LLM $\gM$ for a target downstream task $\gD_{task} = \{q_i,~y_i\}_{i=1}^N$, where $q_i$ denotes the query and $y_i$ is the expected response. The model's performance is enhanced by minimizing the loss between its predictions and the expected outputs.

However, the effectiveness of SFT is heavily dependent on the quality of the downstream task data~\cite{bhatt2024experimental,xia2024less}.
Various factors, including annotation errors, data processing inconsistencies, and model hallucinations, can introduce both random and systematic noise into downstream datasets $\gD_{task}$.
% Existing noise-robust methods~\cite{yuan2024hide,wang2023noise} face significant limitations when applied to LLM fine-tuning scenarios. These methods typically rely on simple majority voting or confidence thresholding, which prove inadequate for handling the complex, open-ended nature of text generation tasks.
Our empirical studies in Section~\ref{sec:experiment} demonstrate that $30\%$ noise in the training data can lead to an $8.9\%$ degradation on downstream tasks. 
Therefore, developing robust mechanisms for noise detection and mitigation during the SFT process, particularly ones that can effectively handle open-ended text generation, is crucial and holds significant practical value for optimizing LLM performance.

\subsection{Problem Definition}

As discussed above, during the fine-tuning of LLMs on downstream tasks, the training data contains both correctly and incorrectly labeled data pairs. Our primary objective is to develop an effective mechanism for identifying these mislabeled instances. Furthermore, we aim to leverage both the model's capabilities and contextual information within the dataset to denoise incorrectly labeled data pairs where possible. Through this process, we seek to construct a refined dataset with reduced noise levels. Ultimately, this curated dataset enables more effective enhancement of LLM performance on downstream tasks.

\section{Methodology}

\subsection{Overview}

Adapting and fine-tuning Large Language Models (LLMs) in real-world scenarios presents significant challenges, particularly due to the presence of noise in downstream task datasets that can compromise model performance. 
Our approach addresses this challenge through a systematic framework comprising noise \textit{detection}-and-\textit{denoising} mechanisms to prevent performance degradation. 

\textit{For noise detection}, we leverage the consensus among multiple expert LLMs and employ a \texttt{Checker} to identify noisy samples. 
\textit{For denoising}, we employ a two-pronged approach: first, we utilize context-enhanced reasoning with clean samples to relabel noisy instances through a \texttt{Review} Agent; second, we implement a perplexity-based data selection mechanism to exclude samples with low confidence scores. 
As demonstrated in Figure~\ref{fig:framework}, this dual-process framework effectively mitigates noise-induced performance deterioration.

\subsection{Noise Detection}

Effective noise identification is crucial for handling noisy data in downstream tasks. In our approach, we leverage collaborative learning among multiple LLMs to uncover potentially noisy samples, enabling a more robust detection mechanism.

Initially, we utilize the base LLM to generate predictions for all data samples:
\begin{equation}\label{eq:LLM-inference}
    \hat{y}_i = \mathcal{M}(q_i)\,,
\end{equation}
where $q_i$ represents the query, $\mathcal{M}$ denotes the LLM and $\hat{y}_i$ is the base prediction.

For internal noise detection, we introduce a reasoning-enhanced LLM that iteratively combines reasoning and reflection processes. This LLM first performs step-by-step reasoning, followed by self-reflection on its reasoning path, and iterates between these two stages to achieve superior reasoning capabilities. For each data sample, this iterative process can be formalized as:
\begin{equation}\label{eq:reasoning-enhanced}
    \hat{y}^{\text{reas}}_i = \gM_{\text{Reas}}\left(q_i,~\gM_{\text{Refl}}\left(\gM_{\text{Reas}}\left(q_i,~\cdots\right)\right)\right) \,,
\end{equation}
where $\hat{y}^{\text{reas}}_i$ represents the final prediction, $\gM_{\text{Reas}}$ and $\gM_{\text{Refl}}$ denote the reasoning and reflection LLMs, respectively, with each reflection stage evaluating and refining the previous reasoning output.

To ensure prediction reliability, we implement a consistency-based \texttt{Checker} mechanism that analyzes multiple prediction sources: the original label ($y_i$), the base LLM prediction ($\hat{y}_i$), and the reasoning-enhanced prediction ($\hat{y}^{\text{reas}}_i$). This mechanism evaluates the agreement among these predictions through a consistency metric:
\begin{equation}\label{eq:checker}
    r_i = \texttt{Checker}(y_i,~\hat{y}_i,~\hat{y}^{\text{reas}}_i) \in \{0,1\} \,,
\end{equation}
where $r_i=1$ indicates high prediction consistency (reliable sample) and $r_i=0$ indicates prediction inconsistency (potentially noisy sample). Based on this consistency evaluation, we partition the dataset into clean samples $\gD_{\text{clean}} = \{(q_i, y_i)| r_i = 1\}$ and potentially noisy samples $\gD_{\text{noise}} = \{(q_i, y_i) | r_i = 0\}$ for subsequent denoising treatment.

\subsection{Data Denoising}

For the potentially noisy dataset $\gD_{\text{noise}}$, we employ a context learning approach for data relabeling, leveraging external knowledge to reduce noise in the data. Specifically, we project queries from both the reliable dataset $\gD_{\text{clean}}$ and the potentially noisy dataset $\gD_{\text{noise}}$ into a shared latent space:
\begin{equation}\label{eq:data-relabeling-encoding}
    h_i = \text{Encoder}(q_i) \in \mathbb{R}^d\,,
\end{equation}
where $h_i$ represents the $d$-dimensional latent representation of query $q_i$ obtained through the encoder network.

During inference, for each noisy sample, we retrieve the $k$ most similar samples from the reliable dataset as context for reasoning:
\begin{equation}\label{eq:context-denoising}
    \hat{y}^{\text{cont}}_i = \gM\left(q_i \;\middle|\; \left\{(q_j,~y_j)\right\}_{j \in \gN_k\left(q_i,~\gD_{\text{clean}}\right)}\right)\,,
\end{equation}
where $\gN_k\left(q_i,~\gD_{\text{clean}}\right)$ denotes the indices of the $k$ most similar samples to $q_i$ in $\gD_{\text{clean}}$ based on their latent representations.

By incorporating external context, we enable the model to generate more reliable responses $\hat{y}^{\text{cont}}_i$. Combined with the previously obtained reasoning-enhanced predictions $\hat{y}^{\text{reas}}_i$, we introduce a \texttt{Review} Agent to evaluate and relabel the data:
\begin{equation}\label{eq:review-agent}
    \tilde{y}_i = \texttt{Review}(q_i,~\hat{y}^{\text{cont}}_i,~\hat{y}^{\text{reas}}_i)\,.
\end{equation}

Through the \texttt{Review} Agent's assessment and synthesis, we obtain the relabeled predictions $\tilde{y}_i$, forming the denoised dataset $\gD_{\text{denoise}}$. However, considering the potential for model errors and uncertainties, we must implement a data selection mechanism for the self-annotated denoised dataset to ensure quality and reliability.

\subsection{Data Selection}

While our denoising process generates a refined dataset $\gD_{\text{denoise}}$ through self-annotation, ensuring the quality of these auto-labeled samples remains crucial. To maintain high data quality and prevent error propagation during subsequent training, we introduce a confidence-based filtering mechanism leveraging entropy metrics. This approach enables us to quantitatively assess the uncertainty in context-enhanced predictions and retain only the most confident samples.

The entropy score for each context-enhanced response is computed as:
\begin{equation}\label{eq:entropy}
    H(\hat{y}^{\text{cont}}_i) = -\frac{1}{N}\sum_{j=1}^N \log p(y_{ij}|q_i, y_{i<j})\,,
\end{equation}
where $p(y_{ij}|q_i, y_{i<j})$ represents the model's prediction probability for the $j$-th token conditioned on the input query and previous tokens, and $N$ denotes the sequence length. Lower entropy scores indicate higher model confidence and more deterministic predictions. Based on these scores, we rank and filter the samples to form our final selected dataset:
\begin{equation}\label{eq:data-selection}
    D_{\text{select}} = \left\{(q_i, \tilde{y}_i) \;\middle|\; \text{rank}\left(H\left(\hat{y}^{\text{cont}}_i\right)\right) \leq \beta |D_{\text{denoise}}|\right\}\,,
\end{equation}
where $\beta$ controls the selection ratio, which defaults to $50\%$ and will be validated in Section~\ref{sec:sen_anal}.

Through this process, we obtain $\gD_{\text{select}}$, which demonstrates reduced noise levels and higher confidence scores through the combined application of denoising relabeling and selective filtering.

\subsection{Summary}

Through the integration of the processes described above, we combine the reliable dataset $\gD_{\text{clean}}$ and the selected denoised dataset $\gD_{\text{select}}$ to form our final fine-tuning dataset $\gD_{\text{ft}}=\gD_{\text{clean}} \cup \gD_{\text{select}}$. Then, we fine-tune the LLM on $\gD_{\text{ft}}$:
\begin{equation}
    \mathcal{M}' = \argmin_{\mathcal{M}} \mathbb{E}_{(q, y) \sim \gD_{\text{ft}}} \left[-\log p_{\mathcal{M}}(y|q)\right]\,,
\end{equation}
where $\mathcal{M}'$ represents the evolved model trained on the noise-reduced downstream task dataset. The complete algorithm is summarized in Algorithm~\ref{alg:algorithm}.

\begin{algorithm}[t]
    \caption{Algorithm of \method{}}
    \label{alg:algorithm}
\begin{flushleft}
\textbf{Require}: Task dataset $\gD_{\text{task}}$, LLM $\gM$;\\
\textbf{Ensure}: Fine-tuned LLM $\gM'$; 
\end{flushleft}
    \begin{algorithmic}[1] 
    \STATE \textit{// Noise Detection} \\
    \STATE Generate base predictions $\hat{y}_i$ using $\gM$
    \STATE Generate reasoning-enhanced predictions $\hat{y}^{\text{reas}}_i$ via iterative reasoning-reflection
    \STATE Use \texttt{Checker} to identify reliable samples
    \STATE Split data into $\gD_{\text{clean}}$ and $\gD_{\text{noise}}$
    \STATE \textit{// Data Denoising}\\
    \FOR{each sample in $\gD_{\text{noise}}$}
        \STATE Generate context-enhanced prediction $\hat{y}^{\text{cont}}_i$
        \STATE Use \texttt{Review} to generate denoised label $\tilde{y}_i$
    \ENDFOR
    \STATE \textit{// Data Selection} \\
    \STATE Calculate entropy scores for denoised samples
    \STATE Select top-$\beta$ confident samples to form $\gD_{\text{select}}$
    \STATE Fine-tune $\gM$ on $\gD_{\text{ft}}$ to obtain $\gM'$
    \end{algorithmic}
\end{algorithm}

\begin{table*}[t]
\centering
\resizebox{\linewidth}{!}{
\setlength{\tabcolsep}{5pt}
\begin{tabular}{@{}l|*{3}{c}|*{3}{c}|*{3}{c}|*{3}{c}|*{3}{c}@{}}
\toprule[1.2pt]
\multirow{2}{*}{\textbf{Method}} & \multicolumn{3}{c|}{\textbf{MMLU}} & \multicolumn{3}{c|}{\textbf{ARC}} & \multicolumn{3}{c|}{\textbf{PubMedQA}} & \multicolumn{3}{c|}{\textbf{Drop}} & \multicolumn{3}{c@{}}{\textbf{FPB}} \\
\cmidrule(lr){2-4} \cmidrule(lr){5-7} \cmidrule(lr){8-10} \cmidrule(lr){11-13} \cmidrule(lr){14-16}
& {30\%} & {50\%} & {70\%} & {30\%} & {50\%} & {70\%} & {30\%} & {50\%} & {70\%} & {30\%} & {50\%} & {70\%} & {30\%} & {50\%} & {70\%} \\
\midrule[1pt]

Vanilla & 65.3 & 65.3 & 65.3 & 82.7 & 82.7 & 82.7 & 72.0 & 72.0 & 72.0 & 87.2 & 87.2 & 87.2 & 75.5 & 75.5 & 75.5 \\
Hermes-3 & 65.5 & 65.5 & 65.5 & 68.7 & 68.7 & 68.7 & 64.8 & 64.8 & 64.8 & 87.1 & 87.1 & 87.1 & 59.4 & 59.4 & 59.4 \\
Tulu-3 & 55.7 & 55.7 & 55.7 & 73.3 & 73.3 & 73.3 & 63.3 & 63.3 & 63.3 & 85.3 & 85.3 & 85.3 & 54.5 & 54.5 & 54.5 \\
SelfLabel & 64.7 & 64.7 & 64.7 & 82.1 & 82.1 & 82.1 & 71.8 & 71.8 & 71.8 & 86.8 & 86.8 & 86.8 & 82.8 & 82.8 & 82.8 \\
SFT & 59.5 & 47.5 & 37.3 & 70.7 & 61.7 & 47.5 & 66.4 & 36.7 & 32.8 & 85.3 & 78.6 & 66.4 & 79.7 & 58.4 & 34.9 \\
NoiseAL & 66.3 & 65.5 & 66.1 & 84.0 & 83.6 & 83.4 & 74.2 & 72.2 & 71.8 & 86.8 & 84.3 & 82.1 & 81.1 & 78.5 & 72.8 \\
SelfRAG & 65.3 & 65.4 & 64.1 & 83.1 & 82.7 & 82.0 & 63.2 & 60.2 & 57.0 & 86.5 & 85.5 & 83.1 & 83.8 & 76.2 & 68.2 \\
SelfSelect & 59.1 & 53.4 & 44.0 & 76.8 & 72.1 & 62.6 & 57.8 & 46.0 & 22.6 & 86.2 & 78.8 & 64.4 & 79.8 & 58.4 & 32.0 \\

\midrule[0.5pt]
\rowcolor{green!10}
\textbf{Ours} & \textbf{68.2} & \textbf{68.0} & \textbf{67.6} & \textbf{84.9} & \textbf{84.7} & \textbf{84.1} & \textbf{75.8} & \textbf{75.6} & \textbf{75.0} & \textbf{90.3} & \textbf{88.5} & \textbf{87.9} & \textbf{84.4} & \textbf{80.5} & \textbf{76.2} \\
\rowcolor{gray!5}
{$\uparrow$ {\scriptsize vs. Vanilla}} & {4.4} & {4.1} & {3.5} & {2.7} & {2.4} & {1.7} & {5.3} & {5.0} & {4.2} & {3.6} & {1.5} & {0.8} & {11.8} & {6.6} & {0.9} \\
\rowcolor{gray!5}
{$\uparrow$ {\scriptsize vs. SFT}} & {14.6} & {43.2} & {81.2} & {20.1} & {37.3} & {77.1} & {14.2} & {106} & {129} & {5.9} & {12.6} & {32.4} & {5.9} & {37.8} & {110} \\
\bottomrule[1.2pt]
\end{tabular}
}
\vspace{-2mm}
\caption{\textbf{Performance comparison} under different noise rates with Llama-3.1 8B. Best results are shown in \textbf{bold}. Numbers in the last two rows show relative improvements (\%).}
\label{tab:model_comparison}
\vspace{-4mm}
\end{table*}

\section{Experiment}\label{sec:experiment}

\subsection{Experiment Setup}

\subsubsection{Datasets}

We conducted comprehensive evaluations on five diverse benchmark datasets: MMLU~\cite{mmlu}, ARC~\cite{clark2018think}, PubMedQA~\cite{jin2019pubmedqa}, Drop, and FPB~\cite{malo2014good}. These datasets span multiple domains and task types: MMLU and ARC evaluate general knowledge across various academic disciplines; PubMedQA tests biomedical reasoning capabilities; Drop assesses numerical reasoning and reading comprehension; and FPB examines financial domain expertise. For each dataset, we constructed experiments with different noise rates (\ie, 30\%, 50\%, and 70\%) to evaluate model performance under different scenarios.

\subsubsection{Backbones and Baselines}

\paratitle{Base Models.} We employed diverse model architectures, including Gemma2-9B~\cite{team2024gemma} and Llama3.1-8B~\cite{dubey2024llama}, along with models of varying parameter sizes such as Llama3.2-3B~\cite{dubey2024llama}.

\paratitle{Baselines.} To comprehensively validate our method's effectiveness, we implemented several baseline approaches: 
(1)~Vanilla: direct model inference; 
(2)~SFT-enhanced solutions utilizing supplementary data to improve LLM performance, including Hermes-3~\cite{teknium2024hermes3technicalreport} and Tulu-3~\cite{lambert2024t}
\footnote{Hermes-3: https://huggingface.co/NousResearch/Hermes-3-Llama-3.1-8B. Tulu-3: https://huggingface.co/allenai/Llama-3.1-Tulu-3-8B-SFT};
(3)~Standard SFT~\cite{hu2021lora} using potentially noisy training data; 
(4)~Denoising approaches, including the state-of-the-art NoiseAL~\cite{yuan2024hide} and LLM-based denoising methods such as SelfLabel and SelfSelect;
(5)~Self-enhancement methods like SelfRAG~\cite{lewis2020retrieval}, which augments inference context using training data. Detailed baseline implementations are provided in the Appendix.

\subsubsection{Implementation Details}

We partitioned each dataset into training and test sets, introducing varying degrees of noise perturbation in the training data. For model fine-tuning, we employed Low-Rank Adaptation~(LoRA)~\cite{hu2021lora} implemented through Llama-factory~\cite{zheng2024llamafactory} across all open-source models. The fine-tuning process was conducted for 2 epochs. We set the $n=4$ and $\theta=50\%$, with further parameter analysis planned for subsequent experiments. The implementation code is available in our anonymous repository. Comprehensive data and training configurations are detailed in the Appendix.

\begin{table*}[t]
\centering
\resizebox{\linewidth}{!}{
\setlength{\tabcolsep}{5pt}
\begin{tabular}{@{}l|*{3}{c}|*{3}{c}|*{3}{c}|*{3}{c}|*{3}{c}@{}}
\toprule[1.2pt]
\multirow{2}{*}{\textbf{Model}} & \multicolumn{3}{c|}{\textbf{MMLU}} & \multicolumn{3}{c|}{\textbf{ARC}} & \multicolumn{3}{c|}{\textbf{PubMedQA}} & \multicolumn{3}{c|}{\textbf{Drop}} & \multicolumn{3}{c}{\textbf{FPB}} \\
\cmidrule(lr){2-4} \cmidrule(lr){5-7} \cmidrule(lr){8-10} \cmidrule(lr){11-13} \cmidrule(lr){14-16}
& 30\% & 50\% & 70\% & 30\% & 50\% & 70\% & 30\% & 50\% & 70\% & 30\% & 50\% & 70\% & 30\% & 50\% & 70\% \\
\midrule[1pt]

\multicolumn{16}{l}{\textit{Llama3.2 3B}} \\
\midrule[0.5pt]
Vanilla & 54.9 & 54.9 & 54.9 & 72.4 & 72.4 & 72.4 & 57.8 & 57.8 & 57.8 & 71.0 & 71.0 & 71.0 & 39.9 & 39.9 & 39.9 \\
SFT & 55.0 & 48.4 & 38.3 & 66.1 & 58.5 & 42.9 & 63.2 & 49.2 & 37.5 & 77.3 & 73.7 & 61.3 & 56.2 & 49.4 & 31.3 \\
\rowcolor{green!10}
\textbf{Ours} & \textbf{58.5} & \textbf{58.2} & \textbf{57.9} & \textbf{74.6} & \textbf{74.3} & \textbf{72.6} & \textbf{68.9} & \textbf{67.9} & \textbf{67.9} & \textbf{78.9} & \textbf{77.6} & \textbf{75.6} & \textbf{66.1} & \textbf{59.4} & \textbf{46.8} \\
\midrule

\multicolumn{16}{l}{\textit{Llama3.1 8B}} \\
\midrule[0.5pt]
Vanilla & 65.3 & 65.3 & 65.3 & 82.7 & 82.7 & 82.7 & 72.0 & 72.0 & 72.0 & 87.2 & 87.2 & 87.2 & 75.5 & 75.5 & 75.5 \\
SFT & 59.5 & 47.5 & 37.3 & 70.7 & 61.7 & 47.5 & 66.4 & 36.7 & 32.8 & 85.3 & 78.6 & 66.4 & 79.7 & 58.4 & 34.9 \\
\rowcolor{green!10}
\textbf{Ours} & \textbf{68.2} & \textbf{68.0} & \textbf{67.6} & \textbf{84.9} & \textbf{84.7} & \textbf{84.1} & \textbf{75.8} & \textbf{75.6} & \textbf{75.0} & \textbf{90.3} & \textbf{88.5} & \textbf{87.9} & \textbf{84.4} & \textbf{80.5} & \textbf{73.2} \\
\midrule

\multicolumn{16}{l}{\textit{Gemma2 9B}} \\
\midrule[0.5pt]
Vanilla & 70.3 & 70.3 & 70.3 & 90.2 & 90.2 & 90.2 & 66.4 & 66.4 & 66.4 & 90.7 & 90.7 & 90.7 & 83.1 & 83.1 & 83.1 \\
SFT & 63.6 & 52.1 & 40.3 & 77.9 & 64.6 & 55.0 & 61.7 & 39.8 & 30.4 & 88.8 & 80.5 & 67.3 & 88.1 & 60.7 & 35.6 \\
\rowcolor{green!10}
\textbf{Ours} & \textbf{72.5} & \textbf{72.1} & \textbf{71.3} & \textbf{91.8} & \textbf{91.5} & \textbf{90.4} & \textbf{70.8} & \textbf{68.8} & \textbf{66.8} & \textbf{91.9} & \textbf{91.8} & \textbf{90.9} & \textbf{91.8} & \textbf{80.8} & \textbf{87.7} \\

\bottomrule[1.2pt]
\end{tabular}
}
\vspace{-2mm}
\caption{\textbf{Performance comparison} across different model architectures and noise rates. Best results for each model are shown in \textbf{bold}.}
\label{tab:comparison_arch_size}
\vspace{-4mm}
\end{table*}
\begin{table}[t]
\centering
\resizebox{\columnwidth}{!}{
\begin{tabular}{rcccccc}
\toprule
\multirow{2}{*}{\textbf{Variant}} & \multicolumn{3}{c}{\textbf{MMLU}} & \multicolumn{3}{c}{\textbf{ARC}} \\
\cmidrule(lr){2-4} \cmidrule(lr){5-7}
& 30\% & 50\% & 70\% & 30\% & 50\% & 70\% \\
\midrule
\rowcolor{gray!10} Llama3.1-8B & & & & & & \\
\rowcolor{gray!10} \quad \textbf{\method{}} & \multirow{-2}{*}{68.2} & \multirow{-2}{*}{68.0} & \multirow{-2}{*}{67.6} & \multirow{-2}{*}{84.9} & \multirow{-2}{*}{84.7} & \multirow{-2}{*}{84.1} \\
\quad \textit{w/o} Selection & 65.7 & 65.1 & 64.6 & 83.2 & 83.0 & 82.8 \\
\quad \textit{w/o} Checker & 65.3 & 65.0 & 64.9 & 82.7 & 82.6 & 82.2 \\
\quad \textit{w/o} Reviewer & 68.0 & 67.7 & 67.1 & 84.5 & 84.3 & 84.0 \\
\quad \textit{w/o} CER & 67.7 & 67.7 & 67.0 & 84.6 & 84.1 & 83.9 \\
\quad \textit{w/o} REL & 67.4 & 67.2 & 66.9 & 84.1 & 83.9 & 83.6 \\
\bottomrule
\end{tabular}
}
\vspace{-2mm}
\caption{\textbf{Ablation study} showing the impact of different noise rates (30\%, 50\%, 70\%) on model variants across MMLU and ARC benchmarks.}
\label{tab:variant_comparison}
\vspace{-4mm}
\end{table}

\subsection{Main Result}

\subsubsection{Comparison with Baselines}

Our comparative experiments with Llama3.1-8B revealed several significant findings. \method{} consistently demonstrated superior performance across all datasets. The experimental results yielded the following key insights.

\paratitle{Noise management is critical in LLM fine-tuning.} The SFT results clearly demonstrate that direct fine-tuning with noisy data substantially degrades model performance, emphasizing the necessity for robust noise detection and removal.

\paratitle{LLMs exhibit limited inherent noise detection capabilities.} SelfSelect's inferior performance compared to SFT indicates that LLMs cannot effectively identify noise, necessitating specialized noise detection and removal mechanisms.

\paratitle{Enhanced SFT approaches lack consistent improvement.} Methods like Tulu-3 and Hermes-3 failed to show uniform performance improvements across downstream tasks, suggesting the need for task-specific LLM adaptation strategies.

\paratitle{Inference enhancement methods show modest gains.} Notably, these approaches achieved some performance improvements despite potential noise in context data, though the improvements were not comparable to our method's results.

\paratitle{Denoising approaches demonstrate mixed results.} While methods such as NoiseAL and SelfLabel show noise resistance and improvements on some datasets, they exhibit degradation on others.

\subsubsection{Comparison across Architectures and Parameter Sizes}
We conducted extensive experiments across multiple model architectures (Llama3.2-3B, Llama3.1-8B, and Gemma2-9B), as shown in Table~\ref{tab:comparison_arch_size}. Our investigation revealed several noteworthy insights:

\paratitle{Larger models are not inherently more robust.} Contrary to common intuition, increased parameter count does not correlate with better noise resistance. In fact, general-purpose large models may be more susceptible to noise during domain-specific fine-tuning due to their lack of domain priors.

\paratitle{Transformation mechanism from general models to domain experts.} While Gemma2-9B showed strong general capabilities, it initially performed worse on domain-specific tasks. However, after \method{}, it effectively adapted to these domains and outperformed Llama3.1-8B, demonstrating the importance of denoising in LLM adaptation.

\paratitle{Critical importance of denoising for smaller models.} Smaller models benefit more significantly from denoising strategies during domain-specific training. Our experiments show that effective denoising mechanisms can substantially mitigate the performance gaps of smaller models in downstream tasks.

\begin{figure}[t] 
    \centering 
    \begin{subfigure}[b]{0.49\columnwidth} 
        \includegraphics[width=\textwidth]{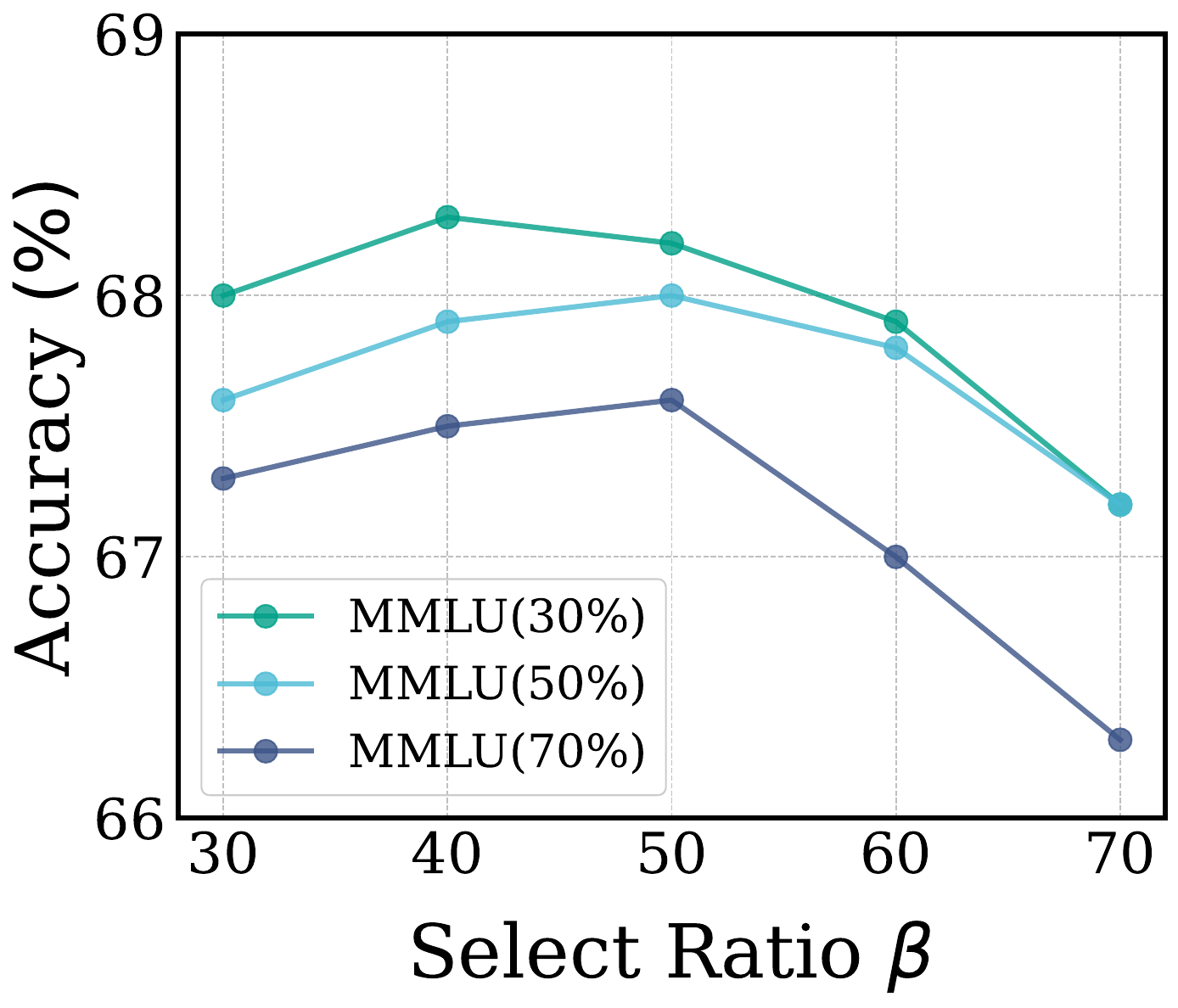}
    \end{subfigure}
    \hfill
    \begin{subfigure}[b]{0.49\columnwidth}
        \includegraphics[width=\textwidth]{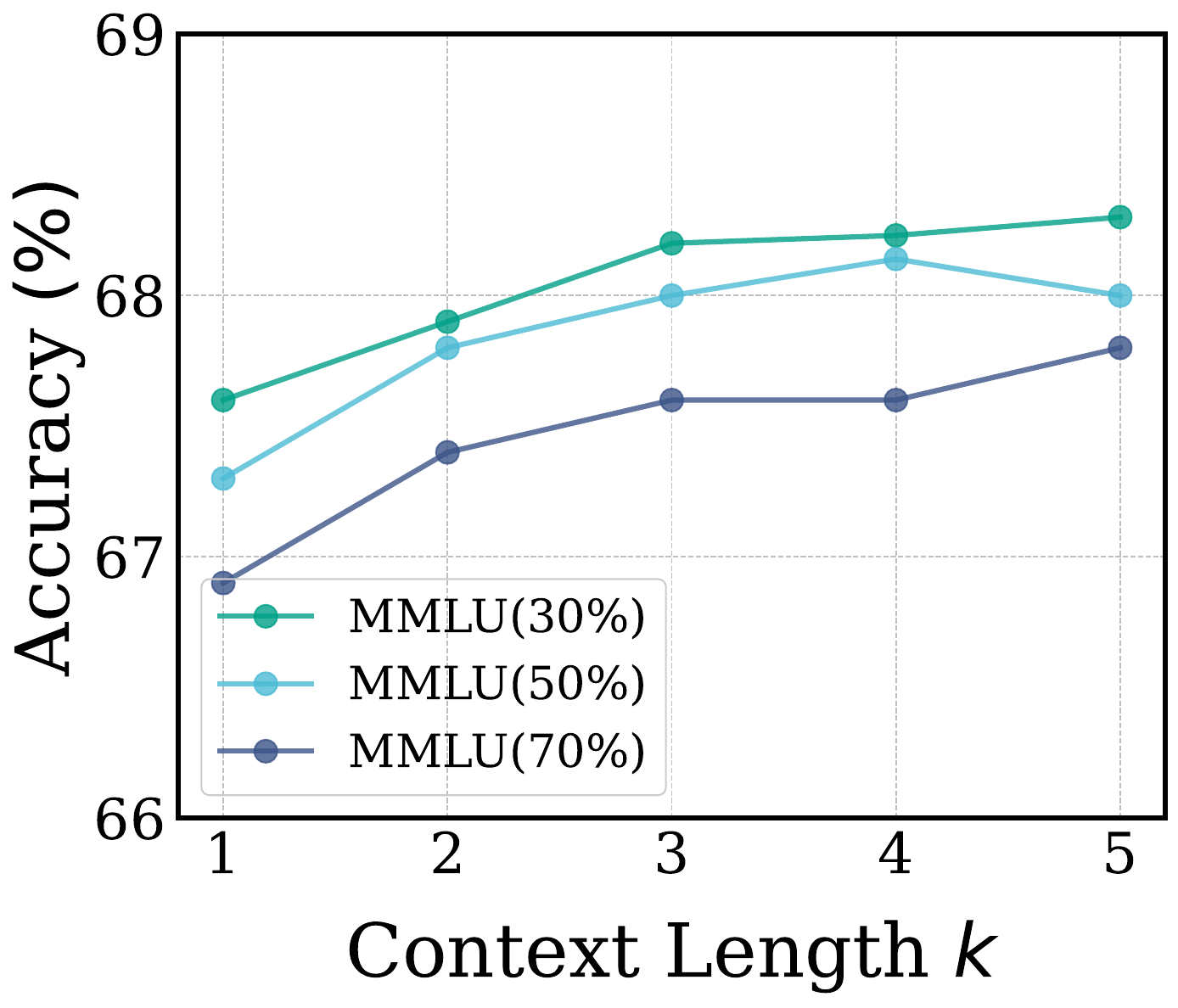} 
    \end{subfigure}
\vspace{-4mm}
\caption{\textbf{Sensitivity analysis} on MMLU under different $\beta$ and $k$ with varying noise levels.}
\label{fig:sensitive} 
\vspace{-4mm}
\end{figure}

\begin{figure*}[t] 
\centering 
\includegraphics[width=\textwidth]{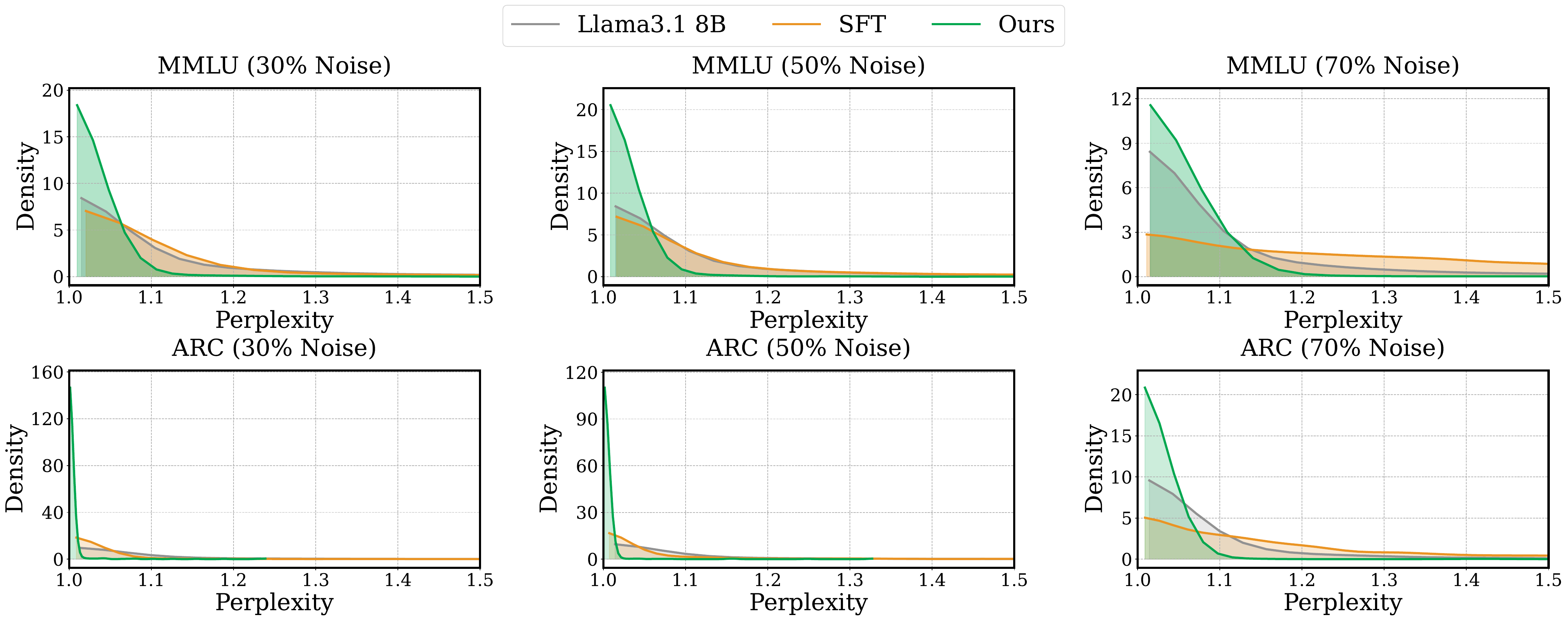}
\vspace{-4mm}
\caption{\textbf{Perplexity analysis} of \method{} on MMLU and ARC with varying noise levels.}
\vspace{-4mm}
\label{fig:ppl} 
\end{figure*}

\begin{figure}[t] 
\centering 
\includegraphics[width=0.8\linewidth]{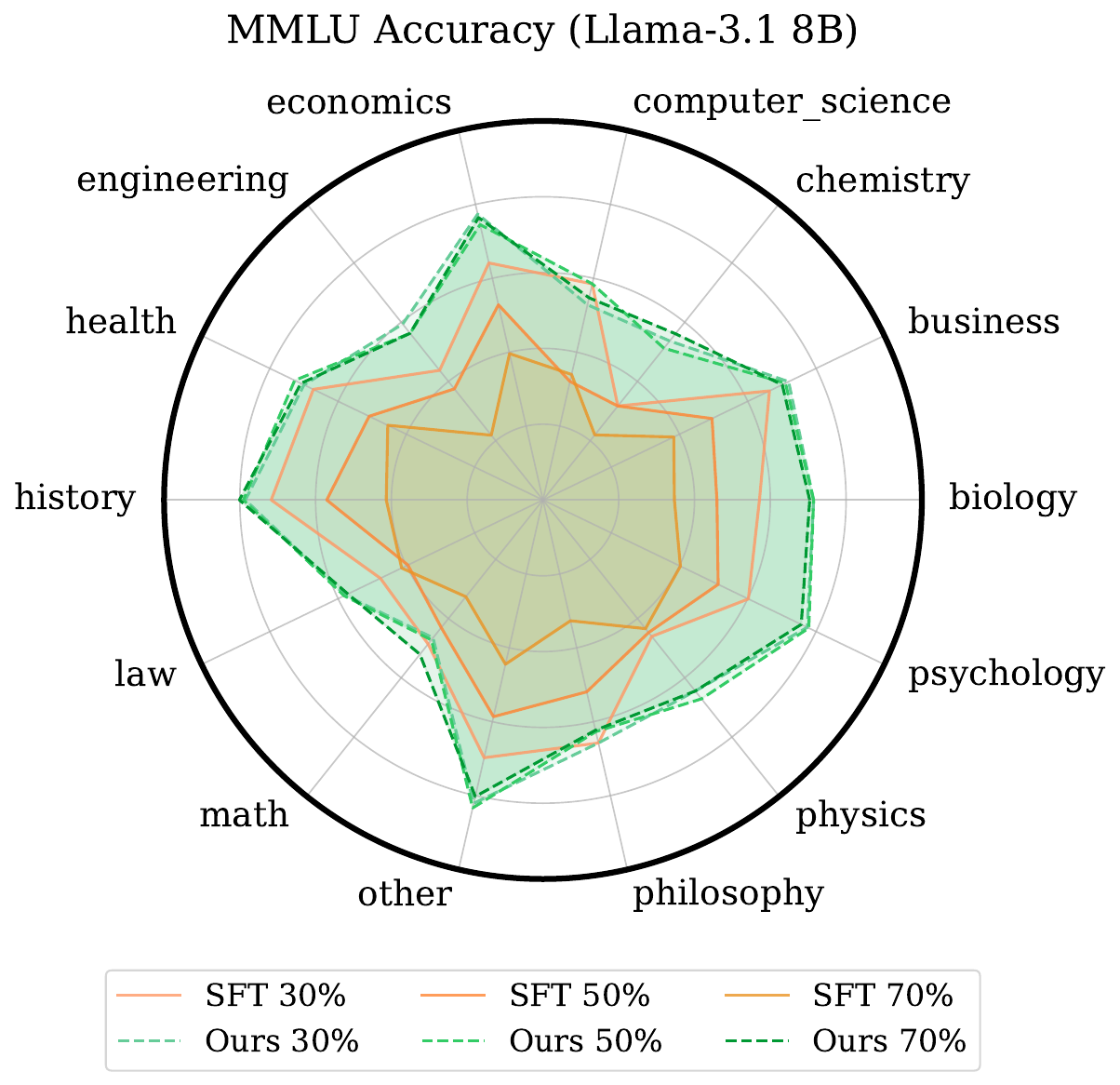}
\vspace{-2mm}
\caption{\textbf{Category-wise performance} of \method{}.}
\vspace{-4mm}
\label{fig:mmlu_radar} 
\end{figure}

\subsection{Analysis and Discussions}

\subsubsection{Ablation Study}

We conducted ablation experiments on RobustFT across different noise levels~(30\%, 50\%, 70\%) using MMLU and ARC datasets. The results reveal several key findings: \textbf{(1)} The complete RobustFT framework consistently achieves optimal performance across all settings, validating its effectiveness. \textbf{(2)} The Selection component proves crucial, as its removal leads to substantial performance drops~(\eg, accuracy decreases from 68.2 to 65.7 on MMLU with 30\% noise). \textbf{(3)} The Checker component significantly contributes to model performance, particularly on the ARC dataset, demonstrating the effectiveness of our multi-model collaborative noise detection. \textbf{(4)} While the Reviewer component shows modest impact, it still contributes to overall data quality. \textbf{(5)} Both Context-Enhanced Relabeing~(CER) and Reasoning-Enhanced LLM~(REL) components prove essential, with their removal leading to notable performance degradation, highlighting the importance of our multi-experts collaborative mechanisms in handling noisy data.

\begin{figure}[t] 
    \centering 
    \includegraphics[width=\linewidth]{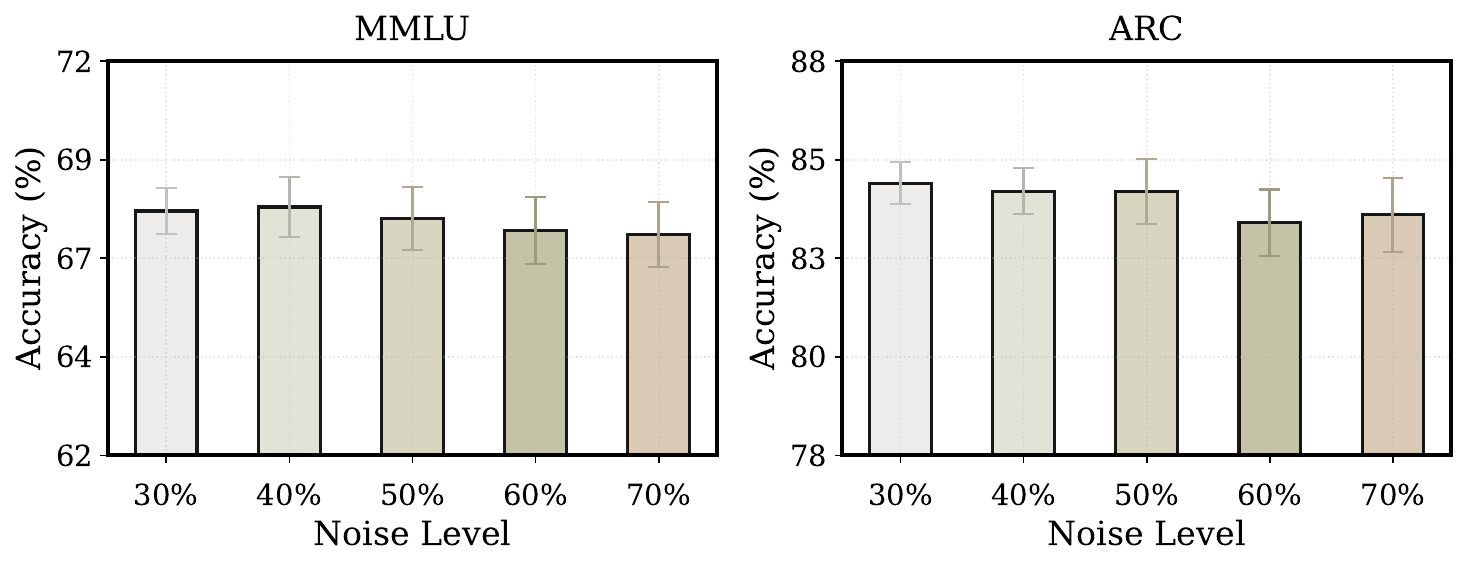}
    \vspace{-4mm}
    \caption{\textbf{Stability analysis} on MMLU and ARC.}
    \vspace{-4mm}
    \label{fig:stable} 
\end{figure}

\subsubsection{Sensitivity Analysis}
\label{sec:sen_anal}

We conducted sensitivity analysis of \method{} on MMLU under different noise levels. As shown in Figure~\ref{fig:sensitive}, we examine the impact of selection ratio $\beta$ and context length $k$. The results show that model performance peaking at $\beta=40-50\%$, with performance degrading significantly beyond this range due to the inclusion of excessive noisy samples. For context length, performance improves with increasing $k$ but plateaus, particularly in the range of $k=3-5$, suggesting that moderate $k$ provide sufficient reasoning support. These findings validate our default parameter choices ($\beta=50\%$, $k=3$) without requiring extensive hyperparameter search, as our primary focus was on demonstrating the framework's overall effectiveness.

\subsubsection{Perplexity Analysis}

We conducted perplexity analysis of the models, as shown in Figure~\ref{fig:ppl}, revealing several key findings: (1) Noise significantly increases perplexity, as evidenced in both SFT and vanilla models. In contrast, \method{} maintains relatively low perplexity levels even with increased noise, demonstrating its robustness. (2) The vanilla model exhibits flatter and more dispersed perplexity distributions, indicating frequent uncertainty in predictions. \method{} effectively concentrates perplexity in lower ranges, suggesting more confident and reliable predictions. (3) The method shows consistency across datasets, with similar perplexity reduction patterns observed on both MMLU and ARC, validating its generalizability across different domains.

\subsubsection{Category-wise Performance Analysis}

We analyzed performance across MMLU categories, as shown in Figure~\ref{fig:mmlu_radar}. \textbf{(1)} The impact of noise varies significantly across different knowledge domains, with knowledge-intensive categories such as History, Healthcare, and Law experiencing more severe performance degradation under noisy conditions. \textbf{(2)} \method{} demonstrates balanced and expanded performance across all categories, achieving comprehensive noise resistance rather than isolated improvements, as evidenced by its smooth and expanded radar plot.

\subsubsection{Stability Analysis}

We evaluated the inference stability of models under different noise conditions, as shown in Figure~\ref{fig:stable}. Specifically, we employed GPT-4o to rephrase the instructions and conducted five independent tests, reporting both mean performance and standard deviation. Results show that \method{} maintains consistent performance, with only minimal variance increase at higher noise rates.

\section{Related Work}

% Data Selection 
% LLM as Judge
% Nosiy Label Learning

\subsection{Noisy Label Learning}
Noisy label learning has been a fundamental challenge in NLP~\cite{yuan2024hide,kim-etal-2024-towards-robust,sun-etal-2023-uncertainty,qi-etal-2023-safer,merdjanovska-etal-2024-noisebench,xu-etal-2024-enhancing,liang2024robust}, primarily focusing on learning from text classification data containing label noise. Existing approaches can be categorized into three main strategies: (1) Sample selection methods~\cite{qiao2022selfmix} that identify clean samples using fixed thresholds, (2) Label correction techniques~\cite{sohn2020fixmatch,zhang2021learning} that rectify original labels based on model predictions, and (3) Consistency regularization approaches~\cite{zhuang2023dygen,northcutt2021confident} that leverage prediction consistency under different perturbations for label refinement.

\paratitle{Challenges in LLM Era.} 
These conventional methods are primarily designed for well-defined scenarios, with finite discrete label spaces, making them less effective for open-ended generation problems. Moreover, LLMs' tendency towards hallucination poses significant challenges in noise detection and correction. To address these limitations, \method{} introduces a novel framework specifically designed for noise-robust downstream fine-tuning of LLMs, moving beyond these constraints.

% 毒性攻击和预防
% 利用毒性有害数据在后训练阶段进攻LLM，及其防御方式收到了广泛关注。目前关于防御机制的研究集中于距离正则化防御，对齐数据混合，提示工程，数据过滤等。与之不同的是，\method{}主要关注于利用检测和重新标注，防止噪音数据的引入导致模型性能下降，而非对毒性数据进行防御。

\subsection{Toxicity Attacks and Defense}
The vulnerability of LLMs to adversarial attacks through toxic and harmful data during post-training stages has garnered significant attention~\cite{huang2024harmful}. Current defense mechanisms primarily focus on several key strategies: distance-based regularization~\cite{mukhoti2023fine,wei2024assessing}, alignment data mixing~\cite{bianchi2023safety}, prompt engineering~\cite{lyu2024keeping}, and data filtering~\cite{choi2024safety}. Different from these methods, \method{} takes a different approach by emphasizing detection and relabeling mechanisms to prevent performance degradation caused by noisy data introduction, rather than specifically defending against toxic content.

\subsection{Self-Evolution and LLM Data Selection}
Recent advances in Large Language Models (LLMs) ~\cite{zhao2023survey} have emphasized the critical role of data quality in Supervised Fine-Tuning (SFT) ~\cite{taori2023stanford,longpre2023flan}. 
Current research primarily explores two approaches: downstream data selection~\cite{bhatt2024experimental,xia2024less,bukharin2023data} and data synthesis~\cite{mukherjee2023orca,chung2024scaling} for improved instruction following. 
To reduce dependence on annotated data, researchers have developed self-evolution methodsthrough self-instruction~\cite{wang2023self} and self-play~\cite{tu2024towards}, enabling models to learn with minimal supervision. 
Additionally, SemiEvol~\cite{luo2024semievol} has demonstrated promising progress by combining a small amount of labeled data with large-scale unlabeled data to enhance LLM performance on downstream tasks.
While existing work focuses on instruction selection~\cite{parkar2024selectllm} and self-training mechanisms~\cite{wang2024self}, \method{} takes a distinct approach by leveraging noisy real-world data for model self-training to enhance downstream performance.

\section{Conclusion}

In this work, we address the practical challenge of handling noisy data in downstream LLM applications, a critical issue that has been unexplored in previous research. We propose a novel noise detection and denoising framework \method{}, which is specifically designed for LLMs. Our approach leverages a multi-expert collaborative mechanism for noise detection, enhanced by a reasoning-enhanced process. Furthermore, we implement context-enhanced reasoning for data relabeling and utilize response entropy for data selection. The effectiveness of \method{} is consistently demonstrated across various datasets and noise scenarios.

% \section*{Limitations}

% Our work has several limitations that warrant acknowledgment. Due to computational resource constraints, we have not evaluated our framework on more larger language models such as GPT-4o and Llama-3.2 70B. Additionally, while our framework demonstrates effectiveness across various benchmark datasets, we have not extensively tested it in more complex domains that require specialized scientific knowledge. These limitations present clear directions for future research: expanding our framework's evaluation to larger-scale LLMs and extending its application to more advanced scientific domains where noise-robust fine-tuning could provide significant value.

% \section*{Ethics Statement}

% Our research adheres to the ACL Code of Ethics. All datasets and language models used in this study are publicly available. The code and related materials will be appropriately released to ensure transparency and reproducibility of our work.

% \clearpage

% \clearpage

\bibliography{custom}

% \clearpage

% \input{7_appendix}

\end{document}